\documentclass[letterpaper, 10 pt, conference]{ieeeconf}  % Comment this line out if you need a4paper

\IEEEoverridecommandlockouts                              % This command is only needed if 
\usepackage[dvipsnames,table,xcdraw, HTML]{xcolor}
\usepackage{graphics} % for pdf, bitmapped graphics files
\usepackage{epsfig} % for postscript graphics files
\usepackage{times} % assumes new font selection scheme installed
\usepackage{amsmath} % assumes amsmath package installed
\usepackage{amssymb}  % assumes amsmath package installed
\usepackage{booktabs}
\usepackage{amsmath}
\usepackage{amssymb}
\usepackage{tabularx}
\usepackage{mathtools}
\usepackage{booktabs}
\usepackage[font=small,labelfont=bf]{caption}

\usepackage[export]{adjustbox}
\usepackage{wrapfig}
\usepackage{enumerate}
\usepackage{pifont}
\usepackage{fontawesome} % for external link symbol
\usepackage{graphicx}
\usepackage{hyperref}
\hypersetup{
    colorlinks=true,
    linkcolor=orange,
    filecolor=magenta,      
    urlcolor=orange,
    citecolor=orange,
}
\usepackage[capitalise, nameinlink]{cleveref}
\makeatletter
\let\NAT@parse\undefined
\makeatother
\usepackage[numbers,sort,compress]{natbib}

\newcommand{\approach}{\text{\textit{Gen2Act}}}

\title{\LARGE \bf
Gen2Act: Human Video Generation in Novel Scenarios enables Generalizable Robot Manipulation
}

\author{Homanga Bharadhwaj$^{1,2}$, Debidatta Dwibedi$^{1}$, Abhinav Gupta$^{2}$, Shubham Tulsiani$^{2}$, Carl Doersch$^{1}$,\\ Ted Xiao$^{1}$, Dhruv Shah$^{1}$, Fei Xia$^{1}$, Dorsa Sadigh$^{1,3}$, Sean Kirmani$^{1}$% <-this % stops a space
\thanks{$^{1}$ Google DeepMind}%
\thanks{$^{2}$ The Robotics Institute, Carnegie Mellon University}%
\thanks{$^{3}$ Computer Science Department, Stanford University}%
\thanks{correspondence to Homanga Bharadhwaj: \texttt{hbharadh@cs.cmu.edu}}
}

\begin{document}

\maketitle
\thispagestyle{empty}
\pagestyle{empty}

%%%%%%%%%%%%%%%%%%%%%%%%%%%%%%%%%%%%%%%%%%%%%%%%%%%%%%%%%%%%%%%%%%%%%%%%%%%%%%%%

%%% TLDR: Learning to predict motion information from web-scale data can enable generalizing robot behaviors to diverse unseen scenarios.
\begin{abstract}
How can robot manipulation policies generalize to novel tasks involving unseen object types and new motions? In this paper, we provide a solution in terms of predicting motion information from web data through human video generation and conditioning a robot policy on the generated video. Instead of attempting to scale robot data collection which is expensive, we show how we can leverage video generation models trained on easily available web data, for enabling generalization. \textit{Our approach~\approach~casts language-conditioned manipulation as zero-shot human video generation followed by execution with a single policy conditioned on the generated video.} To train the policy, we use an order of magnitude less robot interaction data compared to what the video prediction model was trained on. \approach~doesn't require fine-tuning the video model at all and we directly use a pre-trained model for generating human videos. Our results on diverse real-world scenarios show how~\approach~enables manipulating unseen object types and performing novel motions for tasks not present in the robot data. \url{https://homangab.github.io/gen2act/}

\end{abstract}

%%%%%%%%%%%%%%%%%%%%%%%%%%%%%%%%%%%%%%%%%%%%%%%%%%%%%%%%%%%%%%%%%%%%%%%%%%%%%%%%
\section{INTRODUCTION}
%%% HB: new para 1 (done)
To realize the vision of robot manipulators helping us in the humdrum everyday activities of messy living rooms, offices, and kitchens, it is crucial to develop robot policies capable of generalizing to novel tasks in unseen scenarios. In order to be practically useful, it is desirable to not require adapting the policy to new tasks through test-time optimizations and instead being able to directly execute it given a colloquial task specification such as language instructions. Further, such a policy should be able to tackle a broad array of everyday tasks like manipulating articulated objects, pouring, re-orienting objects, wiping tables without the need to collect robot interaction data for every task unlike recent efforts on behavior cloning with robot datasets~\cite{rt1,bharadhwaj2023roboagent,bcz,droid}. This is because collecting large robot datasets that cover the diversity of everyday scenarios is extremely challenging and might be deemed impractical.

\begin{figure}
    \centering
    \vspace*{-0.1cm}\includegraphics[width=\columnwidth]{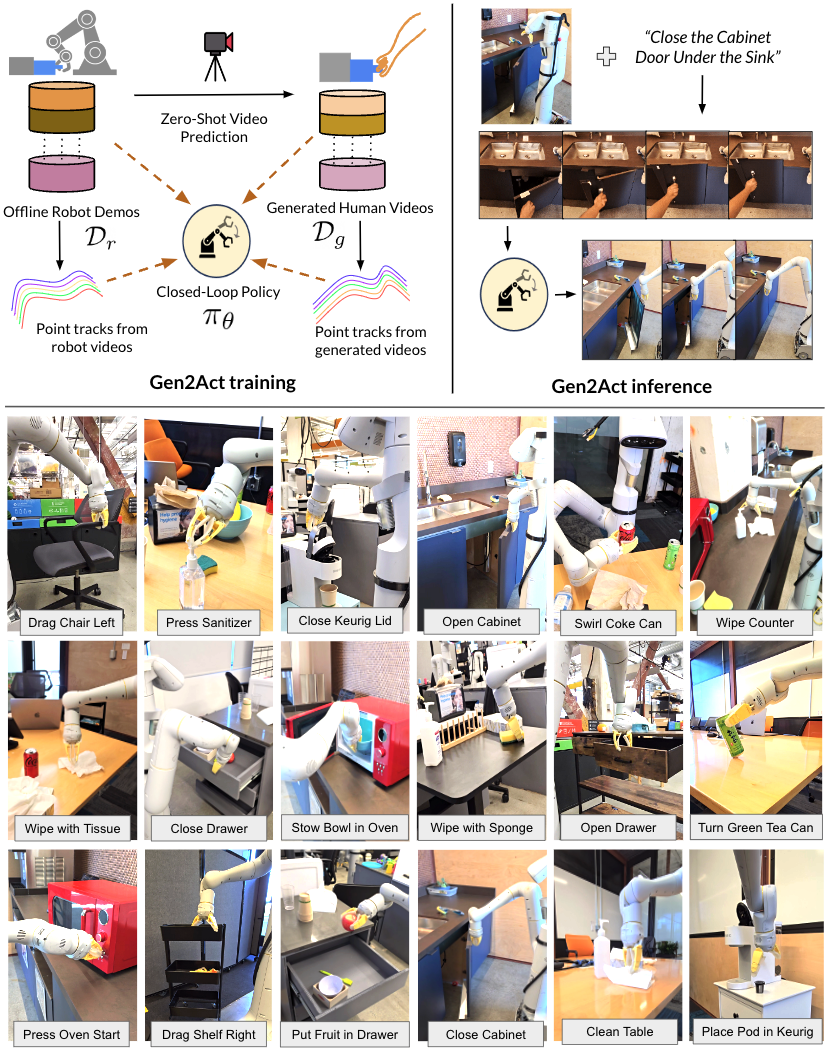}
    \caption{\footnotesize\approach~learns to generate a human video followed by robot policy execution conditioned on the generated video. This enables diverse real-world manipulation in unseen scenarios.}
    \vspace*{-.8cm}
    \label{fig:teaser}
\end{figure}

\begin{figure*}[h!tbp]
    \centering
    \vspace*{-0.1cm}\includegraphics[width=0.95\textwidth]{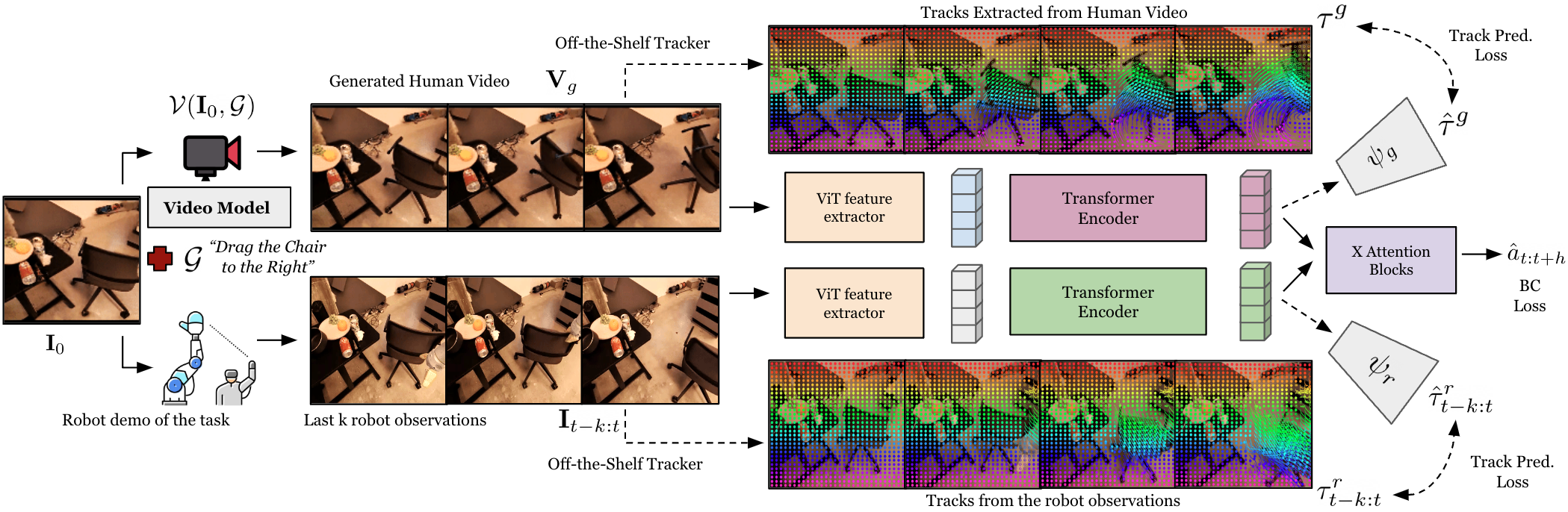}
    \caption{\footnotesize\textbf{Architecture of the translation model of \approach~(closed-loop policy $\pi_\theta$)}. Given an image of a scene $\mathbf{I}_0$ and a language-goal description of the task $\mathcal{G}$, we generate a human video $\mathbf{V}_g$ with a pre-trained video generation model $\mathcal{V}(\mathbf{I}_0,\mathcal{G})$. During training of the policy, we incorporate track prediction from the policy latents as an auxiliary loss in addition to a behavior cloning loss. Dotted pathways show training-specific computations. During inference, we do not require track prediction and only use the video model $\mathcal{V}$ in conjunction with the policy $\pi_\theta(\mathbf{I}_{t-k:t},\mathbf{V}_g)$.}
    \vspace*{-.6cm}
    \label{fig:arch}
\end{figure*}

In order to mitigate issues with purely scaling robotic datasets, a line of recent works have sought to incorporate additional behavioral priors in representation learning by pre-training visual encoders with non-robotic datasets~\cite{r3m,majumdar2023we,pvr2,moco,voltron} and co-training policies with vision-language models~\cite{rt2,dinobota,openvla}. Going beyond abstract representations, other works have learned attributes from web videos more directly informative of motion in the form of predicting goal images~\cite{susie,bharadhwaj2023visual,dallebot}, hand-object mask plans~\cite{bharadhwaj2023towards}, and embodiment-agnostic point tracks~\cite{bharadhwaj2024track2act}. These approaches show promising signs of generalization to tasks unseen in the robot interaction datasets, but training such specific predictive models from web video data requires utilizing other intermediate models for providing ground-truths and thus are hard to scale up. 

Our key insight for enabling generalization in manipulation is to cast motion prediction from web data in the very generic form of zero-shot video prediction. This lets us directly leverage advances in video generation models, by conditioning a robot policy on the generated video for new tasks that are unseen in the robot datasets. We posit that as video generation models get better due to large interest in generative AI~\cite{girdhar2023emu,imagen,kondratyuk2023videopoet} beyond robotics, an approach that relies on learning a policy conditioned on zero-shot video prediction can effectively scale and generalize to increasingly diverse real-world scenarios. For performing a manipulation task in a novel scene, a generated video conditioned on the language description of the task is particularly useful for conveying \textit{what} needs to be done and in capturing motion-centric information of \textit{how} to perform the task that can then be converted to robot actions through a learned policy. Compared to a generated video, a language description or a goal image alone only conveys what the task is. 

% Our key insight for enabling generalization in manipulation is to cast motion prediction from web data in the very generic form of zero-shot video generation, and condition a robotic policy on the generated video. Since video generation models will only get better due to wider community interest beyond robotics~\cite{girdhar2023emu,imagen,kondratyuk2023videopoet}, an approach relying on zero-shot video prediction based policy learning can scale for free and generalize to increasingly diverse real-world scenarios. When a manipulation task to be performed in a novel scene is specified for example in language, a generated video can provide visual cues for both \textit{what} needs to be done and \textit{how} should the relevant object be manipulated. Furthermore, we can obtain explicit motion information from the generated video through off-the-shelf methods like point tracking, tracking hand-object poses etc.  

%% HB: new para 4 (done)
%%% specifics of Gen2Act instantiation
We develop~\approach~by instantiating language-conditioned manipulation as human video generation followed by generated human video to robot translation with a closed-loop policy (\cref{fig:teaser}). We opt for generating human videos as opposed to directly generating robot videos since video generation models are often trained with human data on the web, and they are able to generate human videos zero-shot given a new scene. We then train a translation model that needs some offline robot demonstrations and corresponding generated human videos. We generate these corresponding human videos offline with an off-the-shelf model~\cite{kondratyuk2023videopoet} by conditioning on the first frame of each trajectory (the first frame doesn't have the robot in the scene) and the language description of the task. We instantiate this translation model as a closed loop policy that is conditioned on the history of robot observations in addition to the generated human video so that it can take advantage of the visual cues in the scene and adjust its behavior reactively. 

%% HB: new para 5 (done)
%% training and deployment of Gen2Act
In order to capture motion information beyond that implicitly provided by visual features from the generated video, we extract point tracks from the generated human video and the video of robot observations (through an off-the-shelf tracker~\cite{doersch2024bootstap}) and optimize a track prediction auxiliary loss during training. The aim of this loss function is to ensure that the latent tokens of the closed-loop policy are informative of the motion of points in the scene. We train the policy to optimize the typical behavior cloning loss for action prediction combined with this track prediction loss. For deployment, give a language description of a task to be performed, we generate a human video and run the policy conditioned on this video.

The diverse real-world manipulation results of~\approach~(featured in~\cref{fig:teaser}) demonstrate the broad generalization capabilities enabled by learning to infer motion cues from web video data through zero-shot video generation combined with motion extraction through point track prediction for solving novel manipulation tasks in unseen scenarios. For generalization to novel object types and novel motion types unseen in the robot interaction training data, we show that~\approach~achieves on average $\sim30\%$ higher absolute success rate over the most competitive baseline. Further, we demonstrate how~\approach~can be chained in sequence for performing long-horizon activities like ``making coffee" consisting of several intermediate tasks.

\vspace*{-0.1cm}
\section{Related Works}
\vspace*{-0.1cm}
We discuss prior works in imitation learning with visual observations, learning representations from non-robotic datasets, and approaches for conditional behavior cloning. 

\noindent\textbf{Visual Imitation.} 
Visual imitation is a scalable approach for robotic manipulation~\cite{visual_imitation1,visual_imitation2,visual_imitation3} and end-to-end policy learning more broadly~\cite{visualimitation_broad1,visualimitation_broad2}. 
While early works in multi-task imitation learning collected limited real-world data \cite{mandlekar2018roboturk, jang2022bc},
more recent approaches~\cite{walke2023bridgedata, rt1,kalashnikov2018scalable} collect much larger datasets.
In fact, recent works that have attempted to directly scale this for training large models have required years of expensive data collection~\cite{rt1,rt2,bharadhwaj2023roboagent} and have still been restricted to limited generalization especially with respect to novel object types and novel motions in unseen scenarios.

\noindent\textbf{Visual Representations for Manipulation.} To enable generalization, many recent works propose using pre-trained visual representations trained primarily on non-robot datasets~\cite{ego4d,imagenet}, for learning manipulation policies \cite{r3m, moco,majumdar2023we,vip,majumdar2023we,pvr,pvr2,voltron,wu2023unleashing,yang2024video}. However, they are primarily limited to learning task-specific policies~\cite{r3m, moco, sharma2023lossless, hansen2022pre} as they rely on access to a lot of in-domain robot interaction data. Apart from training visual encoders, a line of works augment existing robot datasets with semantic variations using generative models~\cite{cacti, genaug, rosie,bharadhwaj2023roboagent,chen2024semantically}. While this enables policies to generalize to unseen scenes and become robust to distractors, generalization to unseen object types and motion types still remains a challenge.

\noindent\textbf{Conditional Behavior Cloning.} 
Some prior works train robotic policies conditioned on human videos but require paired in-domain human-robot data~\cite{wang2023mimicplay,smith2019avid,xiong2021learning,vid2robot,wen2023any,gu2023rt} and are not capable of leveraging web data. Others use curated data of human videos to leverage human hand motion information~\cite{qin2021dexmv,shawvideodex} for learning task-specific policies (instead of a single model across generic tasks). Towards learning structure more directly related to manipulation from web videos, some works try to predict visual affordances in the form of where to interact in an image, and local information of how to interact~\cite{mo2021where2act,handsasprobes,bahl2023affordances,liu2022joint,yuan2024general}. While these could serve as good initializations for a robotic policy, they are not sufficient on their own for accomplishing tasks, and so are typically used in conjunction with online learning, requiring several hours of deployment-time training and robot data~\cite{bahl2022human,bahl2023affordances,susie}. Others learn to predict motion from web data more directly in the form of masks of hand and objects in the scene~\cite{bharadhwaj2023towards} and tracks of how arbitrary points in the scene should move~\cite{bharadhwaj2024track2act}, for conditional behavior cloning. However, training such predictive models from web videos requires reliance on intermediate models for providing ground-truth information and are thus hard to scale up broadly.

\vspace*{-0.1cm}
\section{APPROACH}

We develop a language-conditioned robot manipulation system,~\approach~that generalizes to novel tasks in unseen scenarios. To achieve this, we adopt a factorized approach: 1) Given a scene and a task description, using an existing video prediction model generate a video of a human solving the task, 2) Conditioned on the generated human video infer robot actions through a learned human-to-robot translation model that can take advantage of the motion cues in the generated video. We show that this factorized strategy is scalable in leveraging web-scale motion understanding inherent in large video models, for synthesizing \textit{how} the manipulation should happen for a novel task, and utilizing orders of magnitude less robot interaction data for the much simpler task of translation from a generated human video to \textit{what} actions the robot should execute. 

\subsection{Overview and Setup}

Given a scene specified by an image $\mathbf{I}_0$ and a goal $\mathcal{G}$ describing in text the task to be performed, we want a robot manipulation system to execute actions $\mathbf{a}_{1:H}$ for solving the task. To achieve this in unseen scenarios, we learn motion predictive information from web video data in the form of a video prediction model $\mathcal{V}(\mathbf{I}_0,\mathcal{G})$ that zero-shot generates a human video of the task, $\mathbf{V}_g$. In order to translate this generated video to robot actions, we train a closed-loop policy $\pi_\theta(\mathbf{I}_{t-k:t},\mathbf{V}_g)$ conditioned on the video and the last $k$ robot observations, through behavior cloning on a small robot interaction dataset $\mathcal{D}_r$. In order to implicitly encode motion information from $\mathbf{V}_g$ in the policy $\pi_\theta$, we extract point tracks from both $\mathbf{V}_g$ and $\mathbf{I}_{t-k:t}$, respectively $\tau_g$ and $\tau_r$, and incorporate track prediction as an auxiliary loss $\mathcal{L}_\tau$ during training.~\cref{fig:arch} shows an overview of this setup.

\begin{figure}[t]
    \centering
    \vspace*{-0.1cm}\includegraphics[width=\columnwidth]{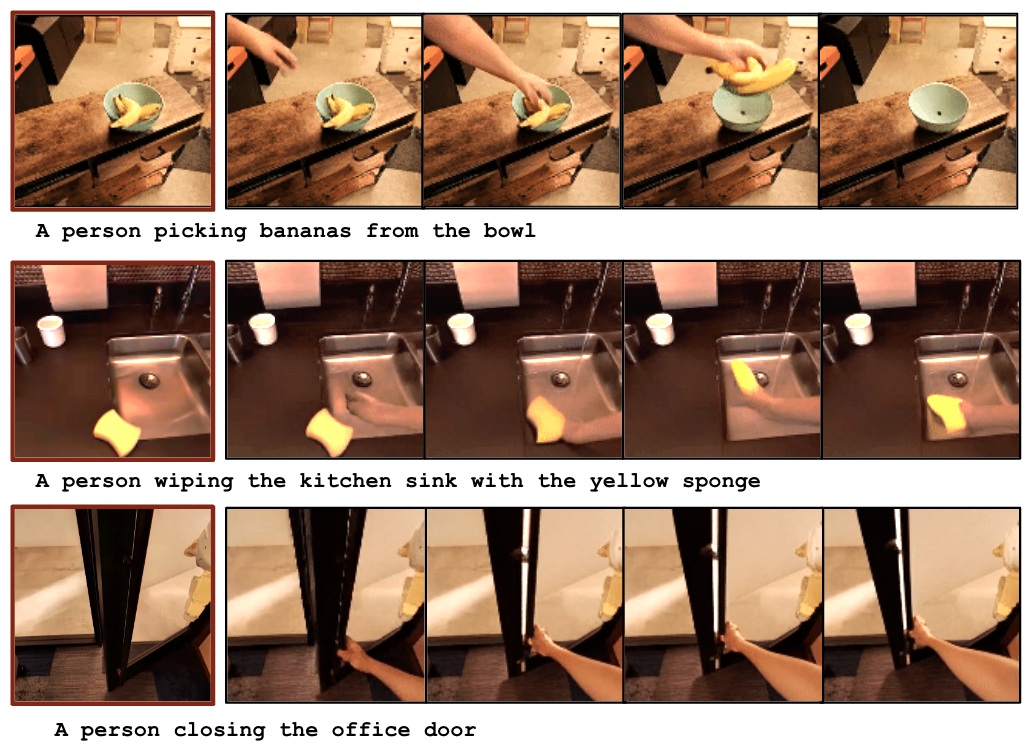}
    \caption{\footnotesize Visualization of zero-shot video generation for different tasks. The blue frame and the language description are input to the video generation model of \approach~and the black frames show sub-sampled frames of the generated video. These results demonstrate the applicability of off-the-shelf video generation models for image+text conditioned video generation that preserves the scene and performs the desired manipulation task.}
        \vspace*{-.7cm}
    \label{fig:qualgenerations}
\end{figure}
\begin{figure*}[t]
    \centering
    \includegraphics[width=\textwidth]{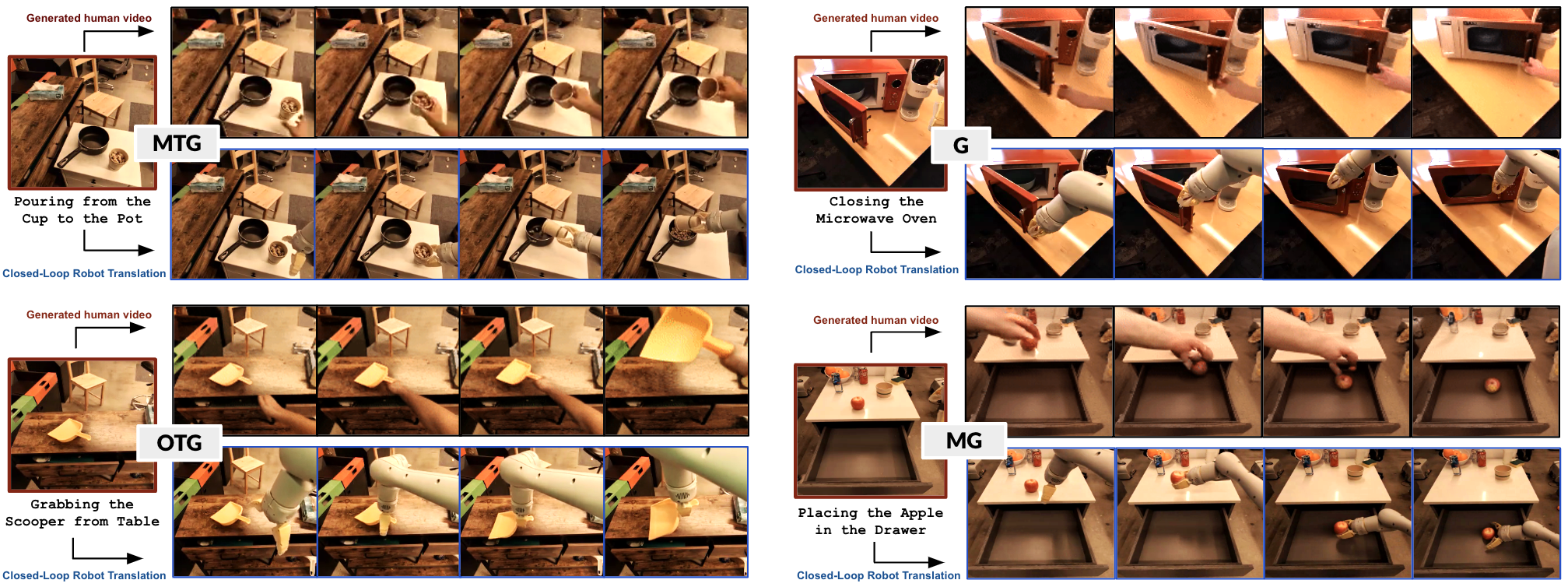}
    \caption{\footnotesize Visualization of the closed-loop policy rollouts (bottom row) conditioned on the generated human videos (top row) for four tasks. The red frame and the language description are input to the video generation model of \approach~. The black frames show sub-sampled frames of the generated video, and the blue frames show robot executions conditioned on the generated video.}
        \vspace*{-.5cm}
    \label{fig:qualrollouts}
\end{figure*}

\subsection{Human Video Generation}
We use an existing video generation model for the task of text+image conditioned video generation. We find that current video generation models are good at generating human videos zero-shot without requiring any fine-tuning or adaptation (some examples in Fig.~\ref{fig:qualgenerations}). Instead of trying to generate robot videos as done by some prior works~\cite{unipy,liang2024dreamitate}, we focus on just human video generation because current video generation models cannot generate robot videos zero-shot and require robot-specific fine-tuning data for achieving this. Such fine-tuning often subtracts the benefits of generalization to novel scenes that is inherent in video generation models trained on web-scale data.

For training, given an offline dataset of robot trajectories $\mathcal{D}_r$ along with language task instructions $\mathcal{G}$, we create a corresponding generated human video dataset $\mathcal{D}_g$ by generating videos conditioned on the first frame of the robot trajectories and the language instruction. This procedure of generating paired datasets $\{\mathcal{D}_r,\mathcal{D}_g\}$ is fully automatic and does not require manually collecting human videos as done by prior works~\cite{rplusx,vid2robot}. We do not require the generated human videos to have any particular structure apart from looking visually realistic, manipulating the relevant objects plausibly, and having minimal camera motion. As seen in the qualitative results in~\cref{fig:qualgenerations}, all of this is achieved zero-shot with a pre-trained video model.

During evaluation, we move the robot to a new scene $\mathbf{I}_0$, specify a task to be performed in language $\mathcal{G}$, and then generate a human video $\mathbf{V}_g = \mathcal{V}(\mathbf{I}_0,\mathcal{G})$ that is fed into the human-to-robot translation policy, described in~\cref{sec:h2rtranslation}. Our approach is not tied to a specific video generative model and as video models become better, this stage of our approach will likely scale upwards. We expect the overall approach to generalize as well since the translation model is tasked with a simpler job of inferring motion cues from the generated human video in novel scenarios, and implicitly converting that to robot actions. As we show through results in~\cref{sec:h2rtranslation} only a small amount of diverse robot trajectories ($\sim 400$) combined with existing offline datasets is enough to train a robust translation model.

\subsection{Generated Human Video to Robot Action Translation}
\label{sec:h2rtranslation}

We instantiate generated human video to robot action translation as a closed loop policy $\pi_\theta$. Given a new scene and a task description, the generated human video provides motion cues for how the manipulation should happen in the scene, and the role of the policy is to leverage relevant information from the generated video, combined with observations in the robot's frame, for interacting in the scene. Instead of attempting to explicitly extract waypoints from the generated video based on heuristics, we adopt a more end-to-end approach that relies on general visual features of the video, and general point tracks extracted from the video. This implicit conditioning on the generated video is helpful in mitigating potential artifacts in the generation and in making the approach more robust to mismatch in the video and the robot's embodiment. Note that we perform human video generation and ground-truth track extraction completely offline for training.

\noindent\textbf{Visual Feature Extraction.}
For each frame in the generated human video $\mathbf{V}_g$ and the robot video $\mathbf{I}_{t-k:k}$, we first extract features, $i_g$ and $i_r$ through a ViT encoder $\chi$. The number of video tokens extracted this way is very large and they are temporally uncorrelated, so we have Transformer encoders $\Phi_g$ and $\Phi_r$ that process the respective video tokens through gated Cross-Attention Layers based on a Perceiver-Resampler architecture~\cite{alayrac2022flamingo} and output a fixed number $N=64$ of tokens. These tokens respectively are $z_g=\Phi_g(i_g)$ and $z_r=\Phi_r(i_r)$. 

In addition to visual features from the generated video, we encode explicit motion information in the human-to-robot translation policy through point track prediction.

\noindent\textbf{Point Track Prediction.}
 We run an off-the-shelf tracking model~\cite{tapir,doersch2024bootstap} on the generated video $\mathbf{V}_g$ to obtain tracks $\tau_g$ of a random set of points in the first frame $P^0$. In order to ensure that the latent embeddings from the generated video $z_g$ can distill motion information in the video, we set up a track prediction task conditioned on the video tokens. For this, we define a track prediction transformer $\psi_g(P^0,i^0_g,z_g)$ to predict tracks $\hat{\tau_g}$ and define an auxiliary loss $||\tau_g - \hat{\tau_g}||_2$ to update tokens $g_e$.  
 
 Similarly, for the current robot video $\mathbf{I}_{t-k:k}$, we set up a similar track prediction auxiliary loss. We run the ground-truth track prediction once over the entire robot observation sequence (again with random points in the first frame $P_0$), but during training, the policy is input a chunk of length $k$ in one pass. So here, the track prediction transformer $\psi_r(P^{t-k},i_{t-k},r^{t-k:t}_e)$ is conditioned on the points in the beginning of the chunk $P_{t-k}$, the image features at that time-step $i^{t-k}$ and the observation tokens for the chunk $z_r$. 

 \noindent\textbf{BC Loss.} For ease of prediction, we discretize the action space such that each dimension has 256 bins. We optimize a Behavior Cloning (BC) objective by minimizing error between the predicted actions $\hat{a}_{t:t+h}$ and the ground-truth $a_{t:t+h}$ through a cross-entropy loss.

In~\approach, we incorporate track prediction as an auxiliary loss during training combined with the BC loss and the track prediction transformer is not used at test-time. This is helpful in reducing test-time computations for efficient deployment. 

\subsection{Deployment}
%% HB: do we need a short section to describe again what deployment looks like (in terms of input/output and intermediate computations?)
For deploying~\approach~to solve a manipulation task, we first generate a human video conditioned on the language description of the task and the image of the scene. We then roll out the generated video conditioned closed-loop policy. For chaining~\approach~to perform long-horizon activities consisting of several tasks, we first use an off-the-shelf LLM (e.g. Gemini) to obtain language descriptions of the different tasks. We chain~\approach~for the task sequence by using the last image of the previous policy rollout as the first frame for generating a human video of the subsequent task. We do this chaining in sequence as opposed to generating all the videos from the first image because the final state of the objects in the scene might be different after the robot execution of an intermediate task.

\section{EXPERIMENTS}

We perform experiments in diverse kitchen, office, and lab scenes, across a wide array of manipulation tasks. Through these experiments we aim to answer the following questions:
\begin{itemize}
    \item Is \approach~able to generate plausible human videos of manipulation in diverse everyday scenes? 
    \item How does \approach~perform in terms of varying levels of generalization with new scenes, objects, and motions?
    \item Can \approach~enable long-horizon manipulation through chaining of the video generation and video-conditioned policy execution?
    \item Can the performance of \approach~for new tasks be improved by co-training with a small amount of additional diverse human tele-operated demonstrations?
\end{itemize}

\subsection{Details of the Evaluation Setup}
\label{sec:genlevels}
Following prior works in language/goal-conditioned policy learning, we quantify success in terms of whether the executed robot trajectory solves the task specified in the instruction, and define success rate over different rollouts for the same task description. We categorize evaluations with respect to different levels of generalization by following the terminology of prior works~\cite{bharadhwaj2024track2act,rt1}:
\begin{itemize}
    \item Mild Generalization (\textbf{MG}): unseen configurations of seen object instances in seen scenes; organic scene variations like lighting and background changes
    \item Standard Generalization (\textbf{G}): unseen object instances in seen/unseen scenes
    \item Object-Type Generalization (\textbf{OTG}): completely unseen object types, in unseen scenes
    \item Motion-Type Generalization (\textbf{MTG}): completely unseen motion types, in unseen scenes
\end{itemize}
Here, seen vs. unseen is defined with respect to the robot interaction data, and the assumption is that the video generation model has seen diverse web data including things that are unseen in the robot data.

\begin{figure*}[t]
    \centering
    \includegraphics[width=\textwidth]{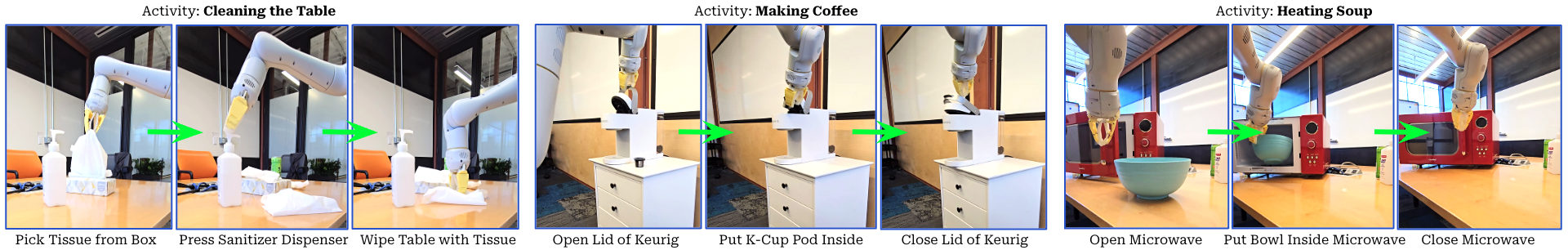}
    \caption{\footnotesize Robot executions for a sequence of tasks. The last frame of the previous execution serves as the conditioning frame for next stage video generation.}
        \vspace*{-.3cm}
    \label{fig:longhorizon}
\end{figure*}
\subsection{Dataset and hardware details}
For video generation, we use an existing video model, VideoPoet~\cite{kondratyuk2023videopoet} by adapting it to condition on square images in addition to language description of tasks. We do not do any fine-tuning of this model for our experiments, and find that it directly generalizes to human video generation in all the robot experiment scenes. 

For robot experiments, we use a mobile manipulator with compliant two finger-grippers, and operate this robot for policy deployment through end-effector control. The arm is attached to the body of the robot on the right. We manually move the robot around across offices, kitchens, and labs and ask it to manipulate different objects in these scenes. We operate the robot for manipulation at a frequency of 3Hz. Before each task, we reset the robot arm to a fixed pre-defined reset position such that the scene is not occluded through the robot's camera.

For training the video-conditioned policy, we use an existing offline dataset of robot demonstrations collected by a prior work~\cite{rt1} and augment this with some paired demonstrations of human videos collected by another prior work~\cite{vid2robot}. In addition, we create pairs of the form $(\texttt{generated\_human\_video}, \texttt{robot\_demo})$ by using the video generation model conditioned on the first frame of the respective robot demo, to generate a corresponding human video. For obtaining tracks on the generated human video and the robot demo, we use an off-the-shelf tracking approach~\cite{tapir,doersch2024bootstap}. Generating human videos, and generating point tracks are done completely offline once and do not induce any additional cost during policy training. 
\begin{table}
\setlength{\tabcolsep}{2.3pt}
\caption{Comparison of success rates for \approach~ with different baselines and an ablated variant for the different levels of generalization as defined in~\cref{sec:genlevels}}
\centering
\begin{tabular}{@{}cccccc@{}}
\toprule
                                 & \begin{tabular}[c]{@{}c@{}}\textbf{Mild}\\\textbf{(MG)}\end{tabular} & \begin{tabular}[c]{@{}c@{}}\textbf{Standard}\\\textbf{(G)}\end{tabular}  & \begin{tabular}[c]{@{}c@{}}\textbf{Obj. Type }\\\textbf{(OTG)}\end{tabular}  & \begin{tabular}[c]{@{}c@{}}\textbf{Motion. Type }\\\textbf{(MTG)}\end{tabular} & \textbf{Avg.} \\ \midrule
\textbf{RT1}                & 68          & 18          & 0           & 0           & 22            \\
\textbf{RT1-GC}                & 75          & 24          & 5           & 0           & 26            \\
\textbf{Vid2Robot}                & \cellcolor{yellow!75} 83          & 38          & 25           & 0           & 37            \\
\textbf{Gen2Act  (w/o track)} & \cellcolor{yellow!75} 83          & \cellcolor{yellow!25} 58          & \cellcolor{yellow!25} 50           & \cellcolor{yellow!25}  5            & \cellcolor{yellow!25} 49            \\
\textbf{Gen2Act}                 & \cellcolor{yellow!75} 83          & \cellcolor{yellow!75} 67          & \cellcolor{yellow!75} 58           & \cellcolor{yellow!75}  
 30           & \cellcolor{yellow!75} 60            \\ \bottomrule
\end{tabular}
\label{tb:resultsoffline}
\end{table}

\subsection{Baselines and Comparisons}
We perform comparisons with baselines and ablations with variants of \approach. In particular, we compare with a language-conditioned policy baseline (\textit{RT1})~\cite{rt1} trained on the same robot data as \approach. We also compare with a video-conditioned policy baseline trained on paired real human and robot videos (\textit{Vid2Robot})~\cite{vid2robot}, a goal-image conditioned policy baseline trained with the same real and generated videos of \approach~ but by conditioning on just the last video frames (i.e. goal image) of the generated human videos (\textit{RT1-GC}). Finally, we consider an ablated variant of \approach~ without the track prediction loss.

\subsection{Analysis of Human Video Generations}
Fig.~\ref{fig:qualgenerations} shows qualitative results for human video generation in diverse scenarios. We can see that the generated videos correspond to plausibly manipulating the scene in the initial image as described by the text instruction. We can see that the respective object in the scene is manipulated while preserving the background and without introducing camera movements and artifacts in the generations. This is exciting because these generations are zero-shot in novel scenarios and can be directly used in a robot’s context to imagine how an unseen object in an unseen scene should be manipulated by a human. 

\subsection{Generalization of \approach~to scenes, objects,  motions}
In this section we compare performance of ~\approach~with baselines and ablated variants for different levels of generalization.~\cref{tb:resultsoffline} shows success rates for tasks averaged across different levels of generalization. We observe that for higher levels of generalization, \approach~achieves much higher success rates indicating that human video generation combined with explicitly extracting motion information from track prediction is helpful in unseen tasks.

\begin{table}
\setlength{\tabcolsep}{0pt}
\centering
\caption{Comparison of success rates for long-horizon activities via chaining of different tasks. We first obtain sub-tasks for activities with an off-the-shelf LLM and then rollout~\approach~in sequence for the different intermediate tasks.}
\begin{tabular}{@{}cccc@{}}
\toprule & \textbf{Activity}
                                 & \textbf{Stages (from Gemini)}                                                                                                                                         & \parbox{3cm}{\centering \textbf{Success} \% \\ Stage 1, Stage 2, Stage 3} \\ \midrule
\multicolumn{2}{c|}{\rotatebox{90}{\parbox{1cm}{\centering\textbf{Stowing}\\\textbf{Apple}}}}  & \begin{tabular}[c]{@{}l@{}}1. Open the Drawer\\ 2. Place Apple in Drawer\\ 3. Close the Drawer\end{tabular}                                         & 80, 60, 60                  \\ \midrule
\multicolumn{2}{c|}{\rotatebox{90}{\parbox{1cm}{\centering \textbf{Making}\\\textbf{Coffee}}}}           & \begin{tabular}[c]{@{}l@{}} 1. Open the Lid \\ 2. Place K-Cup Pod inside\\ 3. Close the Lid\end{tabular}             & 40, 20, 20               \\ \midrule
\multicolumn{2}{c|}{\rotatebox{90}{\parbox{1cm}{\centering \textbf{Cleaning}\\\textbf{Table}}}}        &\hspace{0.5cm} \begin{tabular}[c]{@{}l@{}}1. Pick Tissues from Box\\ 2. Press the Sanitizer Dispenser \\ 3. Wipe the Table with Tissues\end{tabular} & 60, 40, 40                 \\\midrule 
\multicolumn{2}{c|}{\rotatebox{90}{\parbox{1cm}{\centering \textbf{Heating}\\\textbf{Soup}}}}        &\hspace{0.3cm} \begin{tabular}[c]{@{}l@{}}1. Open the Microwave \\2. Put Bowl inside Microwave\\ 3. Close the Microwave\end{tabular} & 40, 20, 20                  \\ \bottomrule
\end{tabular}
        \vspace*{-.3cm}
\label{tb:longhorizon}
\end{table}
\subsection{Chaining~\approach~ for long-horizon manipulation}
We now analyze the feasibility of \approach~for solving a sequence of manipulation tasks through chaining. Table~\ref{tb:longhorizon} shows results for long-horizon activities like ``Making Coffee" that consist of multiple tasks to be performed in sequence. We obtain this sequence of tasks through Gemini~\cite{team2023gemini}, and for each task, condition the video generation on the last image of the scene from the previous execution and execute the policy for the current task conditioned on the generated human video. We repeat this in sequence for all the stages, and report success rates for successful completion upto each stage over 5 trials. \cref{fig:longhorizon} visually illustrates single-take rollouts from four such long-horizon activities.

\subsection{Co-Training with additional teleop demonstrations}
The offline dataset we used for experiments in the previous section had limited coverage over scenes and types of tasks thereby allowing less than $60\%$ success rate of~\approach~for higher levels of generalization (OTG and MTG in~\cref{tb:resultsoffline}). In this section, we perform experiments to understand if adding a small amount of additional \textit{diverse} tele-operated trajectories, for co-training with the existing offline dataset, can help improve generalization. We keep the video generation model fixed as usual. From the results in~\cref{tb:cotraining} we see improved performance of~\approach~ with such co-training. This is exciting because it suggests that with only a small amount of diverse demonstrations, the translation model of~\approach~can be improved to better condition on the generated videos for higher levels of generalization where robot data support is limited.

\begin{table}
\setlength{\tabcolsep}{1.5pt}
\caption{Analysis of co-training with an additional dataset of diverse tele-operated robot demonstrations ($\sim 400$ trajectories).}
\begin{tabular}{@{}cccccc@{}}
\toprule
              Co-Training                                        & \begin{tabular}[c]{@{}c@{}}\textbf{Mild}\\\textbf{(MG)}\end{tabular} & \begin{tabular}[c]{@{}c@{}}\textbf{Standard}\\\textbf{(G)}\end{tabular}  & \begin{tabular}[c]{@{}c@{}}\textbf{Obj. Type }\\\textbf{(OTG)}\end{tabular}  & \begin{tabular}[c]{@{}c@{}}\textbf{Motion. Type }\\\textbf{(MTG)}\end{tabular} & \textbf{Avg.} \\ \midrule
\textbf{Gen2Act (w/o co-train) }     & \cellcolor{yellow!75} 83          & 67          & 58           & 30           & 60            \\ 
\textbf{Gen2Act (w/ co-train)}                 &    \cellcolor{yellow!75}  85       &  \cellcolor{yellow!75}     75      &    \cellcolor{yellow!75}    62      &   \cellcolor{yellow!75}  35         & \cellcolor{yellow!75} 64              \\ \bottomrule
\end{tabular}
        \vspace*{-.5cm}
\label{tb:cotraining}
\end{table}

\subsection{Analysis of Failures}
Here we discuss the type of failures exhibited  by \approach. We observe that for MG and to some extent in G, inaccuracies in video generation are less correlated with failures of the policy. While, for the higher levels of generalization, object type (OTG) and motion type (MTG), if video generation yields implausible videos, then the policy doesn't succeed in performing the tasks. This is also evidence that the policy of~\approach~ is using the generated human video for inferring motion cues while completing a task, and as such when video generation is incorrect in scenarios where robot data support is limited (e.g. in OTG and MTG), the policy fails.

\section{Discussion and Conclusion}

\noindent\textbf{Summary.} In this work, we developed a framework for learning
generalizable robot manipulation by combining zero-shot human video generation from web data with limited robot demonstrations. Broadly,
our work is indicative of how motion predictive models trained on non-robotic datasets
like web videos can be used to used to enable generalization of manipulation policies to unseen scenarios, without requiring collection of robot data for every task. 

\noindent\textbf{Limitations.} Our work focused on zero-shot human video generation combined with point track prediction on the videos as a way for providing motion cues to a robot manipulation system for interacting with unseen objects and performing novel tasks. As such, the capabilities of our system are limited by the current capabilities of video generation models, like inability to generate realistic hands and thereby limited ability to perform very dexterous tasks. 

\noindent\textbf{Future Work.} It would be an interesting direction of future work to explore recovering more dense motion information from the generated videos beyond point tracks, like object meshes for addressing some of the limitations. Another important direction would be to enable reliable long-horizon manipulation by augmenting chaining with learning recovery policies for intermediate failures. 

\section*{Acknowledgements}
We thank Jie Tan for feedback and guidance throughout the project. We are grateful to Peng Xu, Alex Kim, Alexander Herzog, Paul Wohlhart, Alex Irpan, Justice Carbajal, Clayton Tan for help with robot and compute infrastructures. We thank David Ross, Bryan Seybold, Xiuye Gu, and Ozgun Bursalioglu for helpful pointers regarding video generation. We enjoyed discussions with Chen Wang, Jason Ma, Laura Smith, Danny Driess, Soroush Nasiriany, Coline Devin, Keerthana Gopalakrishnan, and Joey Hejna that were helpful for the project. Finally, we thank Jacky Liang and Carolina Parada for feedback on the paper. 
\bibliographystyle{IEEEtran}
\bibliography{example}  % .bib

\newpage
\clearpage
\section*{Appendix}
Here we provide additional details on the method and experiments of \approach.
\subsection{Human Video Generation}
We use a pre-trained VideoPoet model~\cite{kondratyuk2023videopoet} directly without any adaptation or fine-tuning. The input to the model for video generation is a language description of a task (the prompt) and a square-shaped image. By virtue of being trained on diverse large-scale video datasets ($>270M$ videos) we find that this model generalizes well to everyday tasks we develop \approach~for. It can generate realistic and plausible videos of humans manipulating objects, without introducing significant camera motions/artifacts in the generated videos. We ensure that the image of the scene input to the model doesn't have the robot in the frame (the initial reset position of the robot is such that the arm is mostly out of camera view). The language prompt to the model is of the form ``A person \texttt{task-name}, static camera'' e.g. for the task `opening the microwave' the input prompt is ``A person opening the microwave, static camera."

\subsection{Closed-Loop Policy}

For each frame in the generated human video $\mathbf{V}_g$ and the robot video $\mathbf{I}_{t-k:k}$, we first extract features, $i_g$ and $i_r$ through a ViT encoder $\chi$. The number of video tokens extracted this way is very large and they are temporally uncorrelated, so we have Transformer encoders $\Phi_g$ and $\Phi_r$ that process the respective video tokens through gated Cross-Attention Layers based on a Perceiver-Resampler architecture~\cite{alayrac2022flamingo} and output a fixed number $N=64$ of tokens. We use 2 Perceiver-Resampler layers for both the generated video token processing and the robot observation history video processing. These tokens respectively are $z_g=\Phi_g(i_g)$ and $z_r=\Phi_r(i_r)$. During training we sample a fixed sequence of 16 frames from the generated video ensuring that we always sample the first and last frames. For the robot history, we choose the last 8 frames of robot observations. We resize all images to 224x224 dimensions.

 We run an off-the-shelf tracking model~\cite{tapir,doersch2024bootstap} on the generated video $\mathbf{V}_g$ to obtain tracks $\tau_g$ of a random set of points in the first frame $P^0$. In order to ensure that the latent embeddings from the generated video $z_g$ can distill motion information in the video, we set up a track prediction task conditioned on the video tokens. For this, we define a track prediction transformer $\psi_g(P^0,i^0_g,z_g)$ to predict tracks $\hat{\tau_g}$ and define an auxiliary loss $||\tau_g - \hat{\tau_g}||_2$ to update tokens $g_e$. Similarly, for the current robot video $\mathbf{I}_{t-k:k}$, we set up a similar track prediction auxiliary loss. We run the ground-truth track prediction once over the entire robot observation sequence (again with random points in the first frame $P_0$), but during training, the policy is input a chunk of length $k$ in one pass. So here, the track prediction transformer $\psi_r(P^{t-k},i_{t-k},r^{t-k:t}_e)$ is conditioned on the points in the beginning of the chunk $P_{t-k}$, the image features at that time-step $i^{t-k}$ and the observation tokens for the chunk $z_r$. The track prediction transformer has 6 self-attention layers with 8 heads and its role is solely to make the input tokens from generated video / robot observations informative of motion cues. Note that any ground-truth track prediction model can be used for this, and recent advances in point tracking can help improve this step~\cite{cotracker,trackingnew}

For ease of prediction, we discretize the action space such that each dimension has 256 bins. So each action dimension can take values in the range $[0,255]$. The bins are uniformaly distributed within the bounds of each dimension. We predict actions in the end-effector space, and also predict whether to terminate the episode, and whether the gripepr should be open/close. We optimize a Behavior Cloning (BC) objective by minimizing error between the predicted actions $\hat{a}_{t:t+h}$ and the ground-truth $a_{t:t+h}$ through a cross-entropy loss. This discrete action-space for prediction is based on prior works in multi-task imitation learning~\cite{rt1}. 

\subsection{Chaining Gen2Act for Long-Horizon Manipulation}

For deploying~\approach~to solve a manipulation task, we first generate a human video conditioned on the language description of the task and the image of the scene. We then roll out the generated video conditioned closed-loop policy. For chaining~\approach~to perform long-horizon activities consisting of several tasks, we first use an off-the-shelf LLM (e.g. Gemini) to obtain language descriptions of the different tasks. We chain~\approach~for the task sequence by using the last image of the previous policy rollout as the first frame for generating a human video of the subsequent task. We do this chaining in sequence as opposed to generating all the videos from the first image because the final state of the objects in the scene might be different after the robot execution of an intermediate task. This iterative sequencing of video generation after the previous robot execution is made possible by the fast generation capabilities of the Video model we use, VideoPoet (it takes less than 10 seconds to generate a new video after generating the very first video). 

We provide examples of the prompt to Gemini and the subsequent outputs for task decomposition below:\\
\noindent\textit{Cleaning the Table.} Q. What are the steps involved in cleaning a table with sanitizer dispenser and tissues? (write three steps) 

A: Steps for Cleaning a Table with Sanitizer Dispenser and Tissues:

\begin{itemize}
    \item Gather materials: Pick tissues from the box.
    \item Apply sanitizer: Press the sanitizer dispenser onto the tissue.
    \item Wipe: Gently wipe the table surface with the sanitizer-dampened tissues.
\end{itemize}

\noindent\textit{Making Coffee.} Q. What are the steps involved in making coffee with a Keurig? (write first three steps)

A: Steps for Making Coffee with a Keurig:

\begin{itemize}
    \item Open the lid: Open the lid of the Keurig machine.
    \item Insert K-Cup: Place a K-Cup pod into the machine.
    \item Close the lid: Close the lid of the Keurig machine.
\end{itemize}
Videos for these tasks are best viewed in the project website.
\begin{figure*}[t]
    \centering
    \includegraphics[width=0.95\linewidth]{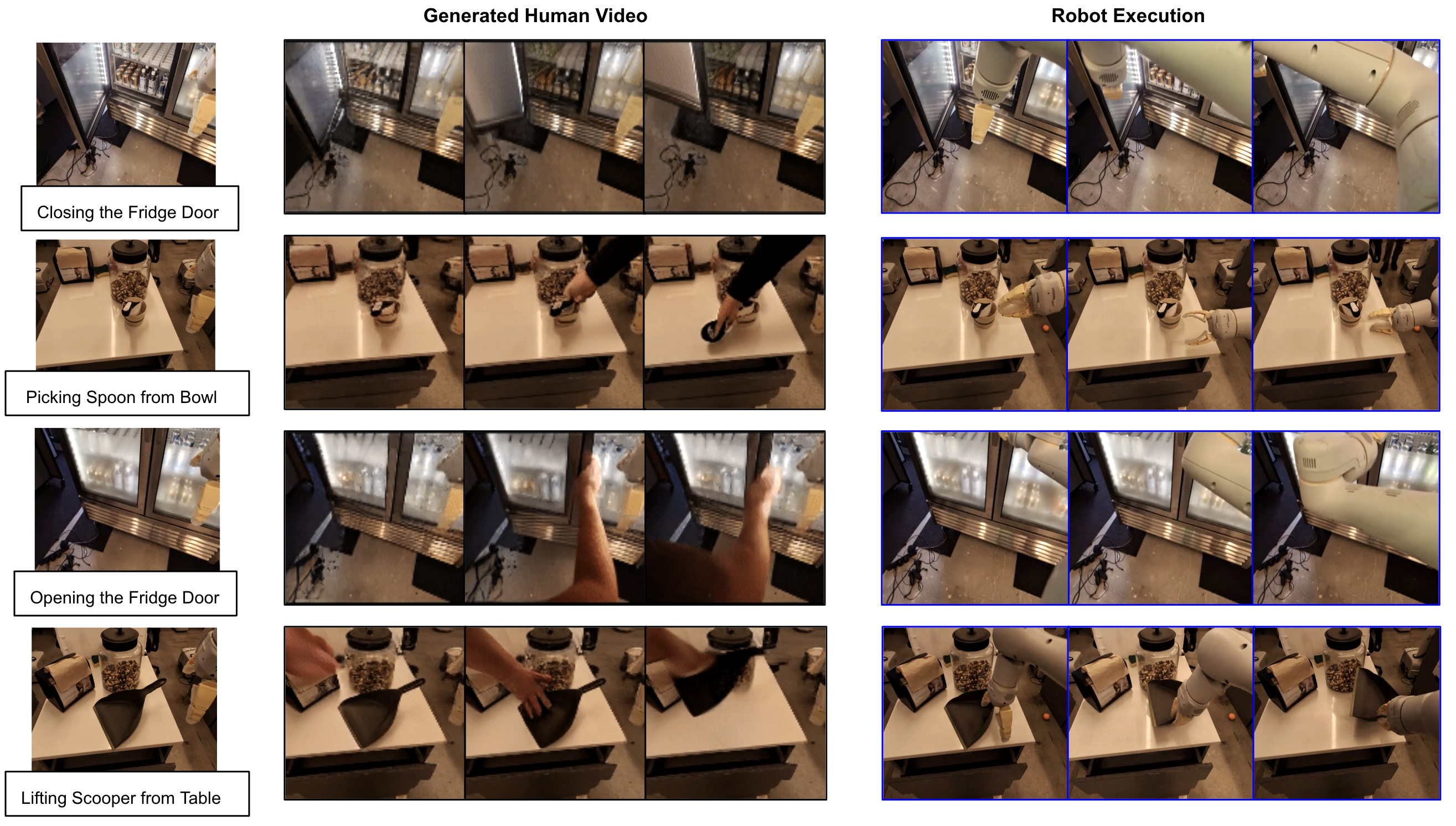}
    \caption{Analysis of failures of~\approach. The tasks here correspond to object type generalization. We can see that most of the failures of robot execution (top 3 rows) are correlated with incorrect video generations. In the last row the video generation is plausible but the execution is incorrect in following the trajectory of the generated video afetr grasping the object.}
    \label{fig:failures}
\end{figure*}
\subsection{Analysis of Failures}
Here we discuss the type of failures exhibited  by \approach. We observe that for MG and to some extent in G, inaccuracies in video generation are less correlated with failures of the policy. While, for the higher levels of generalization, object type (OTG) and motion type (MTG), if video generation yields implausible videos, then the policy doesn't succeed in performing the tasks. This is also evidence that the policy of~\approach~ is using the generated human video for inferring motion cues while completing a task, and as such when video generation is incorrect in scenarios where robot data support is limited (e.g. in OTG and MTG), the policy fails.~\cref{fig:failures} shows some examples of failures of~\approach~in different tasks. Most of the failures are correlated with video generation (first three rows) but generating a video plausibly (fourth row) is not a guarantee of the policy succeeding because there might be issues with grasping the object correctly and following the trajectory of the object post grasp. This indicates potential for future work to explore recovering more dense motion information from the generated videos beyond point tracks, like object meshes for mitigating some of the failures.

\end{document}